\documentclass[runningheads]{llncs}
\usepackage{graphicx}
\usepackage{amsmath,amssymb} %
\usepackage{mdframed}
\usepackage{array}
\usepackage{color}
\usepackage{makecell}
\usepackage{listings}%
\newcolumntype{S}{ >{\centering\arraybackslash} m{.2cm} }
\newcolumntype{I}{ >{\centering\arraybackslash} m{2.5cm} }
\newcolumntype{D}{ >{\centering\arraybackslash} m{.75cm} }
\newcolumntype{T}{ >{\centering\arraybackslash} m{4cm} }

\usepackage[width=122mm,left=12mm,paperwidth=146mm,height=193mm,top=12mm,paperheight=217mm]{geometry}
\begin{document}

\mainmatter

\title{Learning Typographic Style} %

\titlerunning{Learning Typographic Style}

\authorrunning{Shumeet Baluja}

\author{Shumeet Baluja\\
shumeet@google.com}

\institute{Google, Inc.}

\maketitle

\begin{abstract}

Typography is a ubiquitous art form that affects our understanding,
perception, and trust in what we read.  Thousands of different
font-faces have been created with enormous variations in the
characters.  In this paper, we learn the style of a font by analyzing
a small subset of only four letters.  From these four letters, we
learn two tasks.  The first is a discrimination task: given the four
letters and a new candidate letter, does the new letter belong to the
same font?  Second, given the four basis letters, can we generate all of the
other letters with the same characteristics as those in the
basis set?  We use deep neural networks to address both tasks,
quantitatively and qualitatively measure the results in a variety of
novel manners, and present a thorough investigation of the weaknesses
and strengths of the approach.

\end{abstract}

\section{Introduction}

The history of fonts and typography is vast, originating back at least
to fifteenth century Germany with the creation of the movable type
press by Johannes Gutenberg, and the first font ``Blackletter.'' This
was based on the handwriting style of the time, and was used to print
the first books~\cite{tschichold1995,hutchings2012}.  Centuries later,
numerous studies have consistently shown the large impact that fonts
have on not only the readability of text, but also the
comprehensibility and trustability of what is
written~\cite{beymer2008,ha2009,li2010}.

\begin{figure}
\centering
\includegraphics[width=0.7\textwidth]{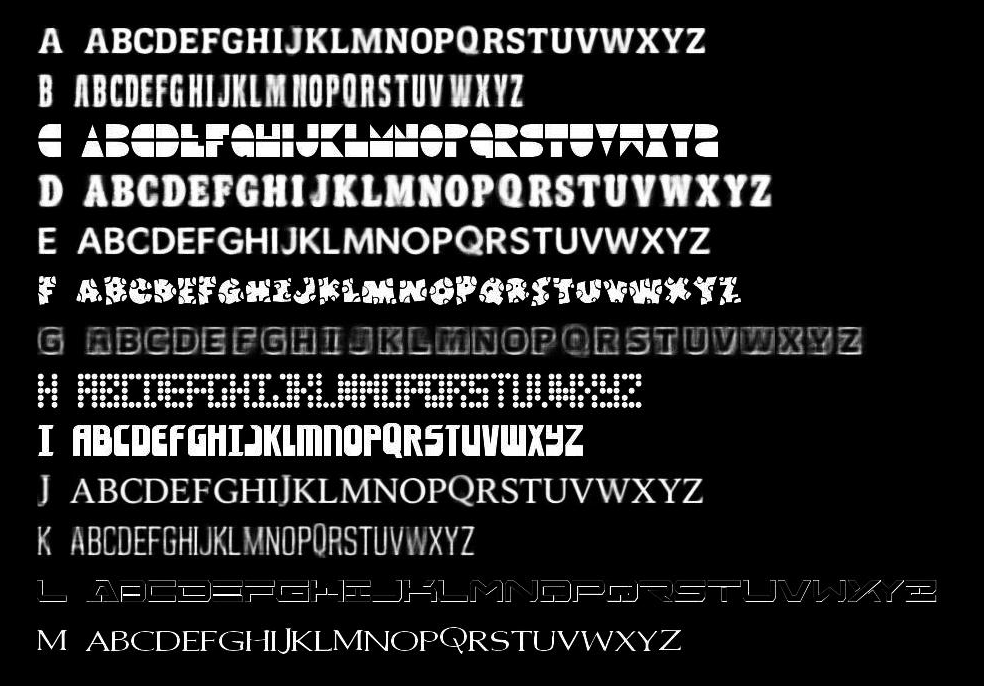}

\caption{Fonts with varying shapes, sizes
  and details.  As a preview, half of these were entirely
  synthesized using our approach. (synthesized rows: A,B,D,E,G,J,K).}
\label{teaser}
\end{figure}

Despite the prevalence of just the few standard fonts used throughout
academic literature, there are innumerable creative, stylized and
unique fonts available. Many have been created by individual designers
as hobbies, or for particular applications such as logos, movies or
print advertisements.  A small sample of a few of the over 10,000
fonts~\cite{10000Fonts} used in this study are shown in
  Figure~\ref{teaser}.

The seminal work of Tenenbaum and Freeman~\cite{tenenbaum2000} towards
separating style from content was applied to letter generation.  Our
motivation and goals are similar to theirs: we hope that a learner can
exploit the structure in samples of related content to extract
representations necessary for modeling style.  The end-goal is to
perform tasks, such as synthesis and analysis, on new styles that have
never been encountered.  In contrast to~\cite{tenenbaum2000}, we do
not attempt to \emph{explicitly} model style and content separately;
rather, through training a learning model (a deep neural network) to
reproduce style, content and style are \emph{implicitly}
distinguished. We also extend their work in four directions.  First,
we demonstrate that a very small subset of characters is required to
effectively learn both discriminative and generative models for
representing typographic style; we use only 4 instead of the 54
alpha-numeric characters used previously.  Second, with these 4
letters, we learn individual-letter and combined-letter models capable
of generating all the remaining letters.  Third, we broaden the
results to thousands of unseen fonts encompassing a far more expansive
set of styles than seen previously. Finally, we present novel methods
for quantitatively analyzing the generated results that are applicable
to any image-creation task.

Fonts are particularly well suited for the examination of style: they
provide diversity along many axes: shape, size, stroke weight, slant,
texture, and serif details --- all within a constrained environment in
which the stylistic elements can be readily distinguished from
content.  Additionally, unlike many image generation tasks, this task
has the enormous benefit of readily available ground-truth data,
thereby allowing quantitative measurement of performance.  Recently,
growing attention has been devoted to style, not only in terms of
fonts~\cite{torres2016,upchurch2016,bernhardsson2016}, but in perceptual shape similarity for
architecture and rigid objects~\cite{lun2015elements}, computer
graphics~\cite{willats2005defining}, cursive
text~\cite{graves2013sequence}, 
photographs~\cite{karayev2013,gygli2013,khosla2014}, 
artwork~\cite{gatys2015}, and music~\cite{aucouturier2003}.

The primary goal of this paper can be succinctly stated as follows:
given a small set of letters (the basis set) of a particular font, can
we generate the remaining letters in the same style?  Before we can
address this question, we need to ensure that there is enough
information in a small basis set to ascertain \emph{style}.  This is
the subject of the next section.

\section {A Discriminative Task: Same Font or Not?}
\label{discriminator}

By addressing the question of whether a set of basis-letters contain
enough information to extract ``style'', we immediately work towards
identifying sources of potential difficulties for the overarching goal
of generating characters -- \emph{i.e.} will the generative process
have enough information to extract from the basis letters to
successfully complete the task?  Further, the networks created here
will be a vital component in validating the final generated results
(Section~\ref{validating}).

\begin{figure}
\centering
  \begin{minipage}[c]{0.49\textwidth}
\centering
\includegraphics[width=0.4\textwidth]{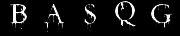}
\includegraphics[width=0.4\textwidth]{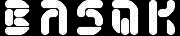}
\\
\includegraphics[width=0.4\textwidth]{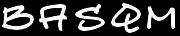}
\includegraphics[width=0.4\textwidth]{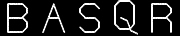}
\\
\includegraphics[width=0.4\textwidth]{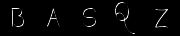}
\\

  \end{minipage}
  \begin{minipage}[c]{0.49\textwidth}
\centering
\includegraphics[width=0.4\textwidth]{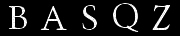}
\includegraphics[width=0.4\textwidth]{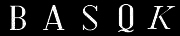}
\\
\includegraphics[width=0.4\textwidth]{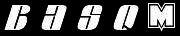}
\includegraphics[width=0.4\textwidth]{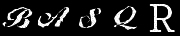}
\\
\includegraphics[width=0.4\textwidth]{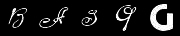}
\\
  \end{minipage}

\caption{Positive (Left) and Negative (Right) samples for the
  classification task. In the left group, the fifth letter is a member
  of the same font.  On the right, the fifth letter is not.  Note that
  some cases on the right are easy.  However, the top two are more
  challenging, requiring the slant and subtle edge weights to be
  analyzed, respectively.}

\label{classify}
\end{figure}

The discriminative task is as follows: Given a set of four basis
letters\footnote{B,A,S,Q comprised the basis set throughout this
  study.  They were chosen because they contained a diverse set of
  edges, angles and curves that can be rearranged/analyzed to reveal
  hints for composing many of the remaining letters.}, \emph{e.g.}
B,A,S,Q, in font-A and another letter $\Phi$, can we train a
classifier to correctly identify whether $\Phi$ is generated in
font-A's style?  See Figure~\ref{classify}.  Next, we describe how the
data is pre-processed and the neural networks used to address the
task.

\subsection {Data Specifics}

Due to the large variation in the font-faces examined, normalization
of the font images was vital, though minimal.  Each letter was input
to the network as a 36$\times$36 gray-scale pixel image. In total, 5
letters were used as inputs: the first 4 were the basis letters (all
from the same font), and the fifth letter was to be categorized. The
output of the network was a single binary value, indicating whether
the fifth letter belonged to the same font as the other 4.

All the fonts used in this study were True-Type-Font format (TTF),
which allows for scaling. Due to the stylistic variations present in
the data, the ratio of widths to heights of different font faces vary
dramatically.  The size of the characters within the 36$\times$36
image is set as follows.  For each font, all 26 capital letters are
generated with the same point-setting until the bounding box of
\emph{any one} of the 26 characters reaches the 36$\times$36 limit in
either dimension.  All 26 characters are then generated with that
point setting; this ensures that if the font style specifies having a
large 'R' while having a little 'L', that stylistic decision is
preserved in the training example and that the largest letter fits
within the 36$\times$36 bounding box.  Note, that this sometimes
creates non-standard positioning of letters when seen together.  For
example, the character 'Q' sometimes has a
\emph{descender}~\cite{bowey2009} that extends below the font's
baseline; this will appear raised up in the training examples (see
Figure~\ref{classify}).

A training set of approximately 200,000 examples was created from
10,000 randomly chosen unique fonts. The training set was composed of
100,000 positive examples (5th letter was the same font) and 100,000
negative examples (5th letter was a different font).  When generating
the examples, each character was randomly perturbed from the center
location by up to $\pm$2 pixels to introduce variation (this also
allowed us to create enough unique positive examples).  A disjoint
testing set with 1,000 font families not present in the training set
was created in exactly the same manner.  For both training and
testing, negative examples were chosen by randomly pairing different
font-families\footnote{To create negative examples, two fonts from the
  same \emph{font-family}, i.e. \emph{Times-Roman-Bold} and
  \emph{Times-Roman} were never used in the same negative example.
  Because the hierarchy of fonts is not \emph{apriori} known, font
  families were estimated by a simple lexicographic analysis of the
  font names.  Importantly, it should be noted that this did not
  preclude fonts that appear visually nearly indistinguishable (but
  with dissimilar names) from being used as negative examples.}.

\subsection {Network Architectures}

Building on the successes of deep neural networks for image
recognition tasks (see ~\cite{imagenet2015} for a recent summary with
ImageNet), we explore a variety of neural network architectures for
this task.  Numerous experiments were conducted to discover effective
network architectures to use.  Of the over 60 architectures tried, the
top performing seven were selected.  Though a full description of the
experiments is beyond the scope of this paper, a few of the general
principles found are provided here to help guide future studies.

\begin{enumerate}

\item \emph {Treating inputs as a single large image or as 5
  individual images:} We compared using the full
  36$\times$(36$\times$5) pixel image as a single input vs. creating
  individual 'towers' for each of the 5 characters
  (see~\cite{karpathy2014,ciresan2012,szegedy2015going,wang2015} for a
  review of column/tower architectures).  As a single image, it may be
  easier to capture the relative sizes of characters.  As multiple
  images, individualized character transformations can be created with
  potentially cleaner derivatives.  Through numerous experiments, the
  results consistently favored using individual towers, both in terms
  of training time and final accuracy.  See Figure~\ref{twoTowers}.
\begin{figure}
\centering

    \includegraphics[height=1.5in]{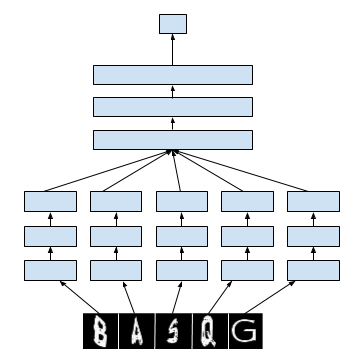}
    \includegraphics[height=1.5in]{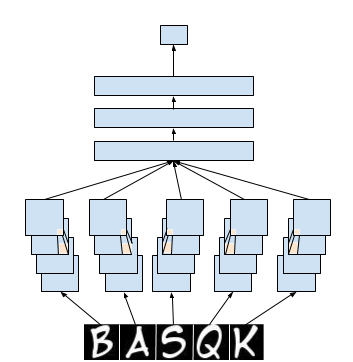}
~~~~~~~
    \includegraphics[height=1.5in]{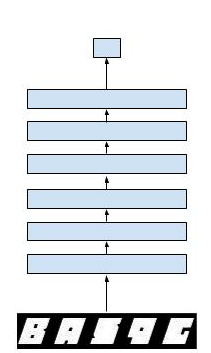}
\caption{Left \& Middle: Two tower-based network architectures. Left:
  Layers within the towers are fully connected.  Middle: Layers within
  the towers are convolutions.  Right: Fully connected net that
  treated the 5 character inputs as a single image (not shown:
  experiments with single-image-input networks that employ
  convolutions were also conducted).  The figure also shows three
  different fonts used in training all the networks.}
\label{twoTowers}

\end{figure}

\item \emph {Network Depth:} In contrast to the trend of increasing
  network depth, deeper nets did not lead to improved performance.
  Unlike in general object detection tasks in which deep neural
  networks are able to exploit the property that images can be
  thoughts of as compositional hierarchies~\cite{lecun2015}, such deep
  hierarchies may not be present for this task.  For this study, two
  sets of convolution layers were used with fully connected layers
  prior to the final output (more details in
  Section~\ref{classificiationResults}).  Additional depth did not
  increase performance.

\item \emph {RELU or Logistic?} A variety of activation functions were
  attempted, including exclusively logistic activations throughout the
  network (recall that the networks were not as deep as is often used
  in image-recognition tasks, a condition often cited for using
  Rectified Linear (RELU) activations~\cite{zeiler2013rectified}).
  Except for the final logistic-output unit, the rest of the
  activations throughout the network worked best as RELU.  Though the
  performance difference was not consistent across trials, RELU units
  trained faster.

\item \emph {Convolutions vs. Fully Connected Layers:} Convolution
  layers are frequently employed for two purposes: achieving
  translation invariance and free parameter reduction, especially when
  used in conjunction with pooling layers.  For this task, translation
  invariance is not as crucial as in general object detection tasks,
  as the object of interest (the character) is centered.  Further,
  this task worked well across a number of network sizes and
  free-parameter ranges.  Networks that employed convolutions as well
  as those that used fully connected layers exclusively performed
  equally well.

\end{enumerate}

\subsection {Individual and Ensembles Network Results}
\label{classificiationResults}

As mentioned in the previous section, over 60 architectures were
experimented with for this task.  Gradient descent with momentum
(SGD+Momentum) was used to train the networks and no domain-specific
prenormalization of the weights was necessary.  From the 60
architectures, seven were chosen (see Table~\ref{architectures}). All
are based on the tower architecture (Figure~\ref{twoTowers}).  The
performance on the discrimination task was measured on an independent
test set --- the 4-basis letters and $\Phi$ were drawn from fonts that
were not used for training.  The results were consistent across a wide
variety of free parameters; in the seven networks, the number of
parameters varied by a factor of 18$\times$, with similar performance.

\begin{table}
 \centering
 \caption {Seven top performing architectures found from the 60 networks tried. }
 \scriptsize
 \tiny
 \label {architectures}
 \begin {tabular} {  c | c |c | c | c}
   network & \makecell {Description of \\ hidden layers \\in each tower} & \makecell {Fully
     Connected \\hidden units in \\aggregation layers} &
   \makecell{Total free\\ parameters} & performance \\
   \hline
   \hline
     1 & 1 Fully Connected (50 units) & 2 (250,200) & 437,602  &   \scriptsize{90.3\%} \\
     2 & 3 Fully Connected (50,50,50 units)& 2 (250,200) & 463,102 & \scriptsize{89.2\%} \\
     3 & 3 Fully Connected (100,100,100 units) & 2 (250, 200) & 925,352 & \scriptsize{88.2\%}
     \\ 
     4 & 3 Fully Connected (100,100,100 units) & 2 (50, 50) & 777,202 & \scriptsize{89.3\%} \\ 
     5 & \makecell{2 Conv Paths, 2 Deep\\(3x3 $\rightarrow$ 3x3, 4x4 $\rightarrow$ 3x3)} & 2 (50, 50) & 216,952 & \scriptsize{90.7\%} \\ 
     6 & \makecell{2 Conv Paths, 2 Deep\\(3x3 $\rightarrow$ 3x3, 4x4 $\rightarrow$ 3x3)} & 4 (50, 50, 50, 50) & 222,052 &
     \scriptsize{91.1\%} \\ 
     7 & \makecell{2 Conv Paths, 2 Deep\\(3x3 $\rightarrow$ 3x3, 4x4 $\rightarrow$ 3x3)}& 4 (10, 10, 10, 10) &
     52,612 & \scriptsize{90.0\%} \\ 
   \hline
     \textbf{Voting Ensemble} & & & & \footnotesize{92.1\%} \\
  \hline
 \end {tabular}
\end{table}

\begin{figure}
\centering
    \includegraphics[height=2in]{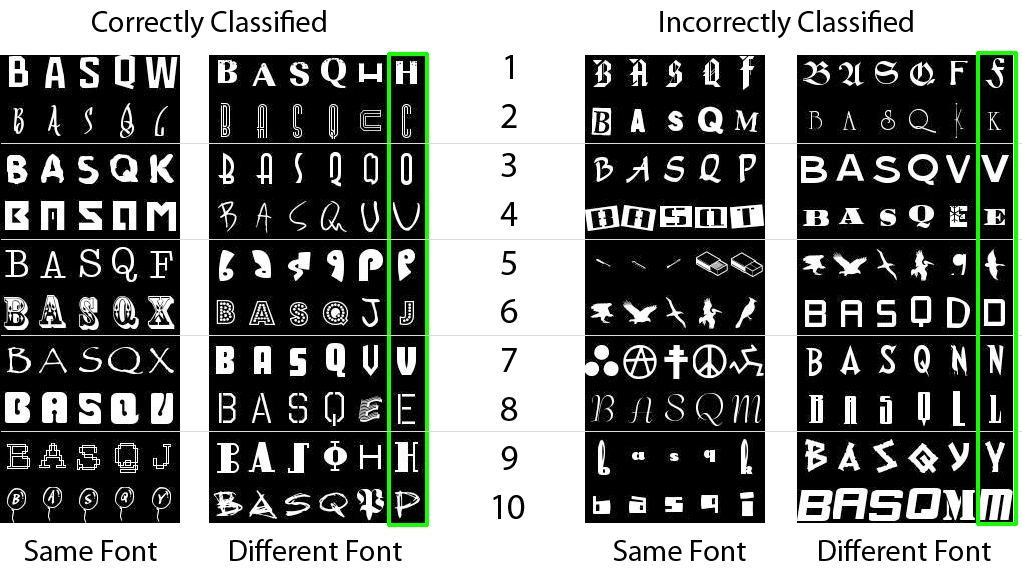}\\
\caption{Same Font or Not?  10 binary classification results.  In
  columns 2 \& 4, which show examples in which the 5th character is
  taken from a different font than the basis set, the corresponding
  \emph{actual} letter from the basis set is also given.  This sixth
  character, ``the ground-truth,'' is not used in training/testing; it is
  only shown here for comparison.}
\label{classificationResults}
\end{figure}

It is illuminating to visualize the types of errors that the networks
made.  Figure~\ref{classificationResults} provides examples that were
both correctly (Left) and incorrectly (Right) classified by all 7
networks.  In the leftmost column, examples in which the 5th character
was correctly recognized as belonging to the same font are shown.  In
the second column, correctly recognized negative examples (the 5th
letter was a different font) are given. Note that the 6th letter, the
``ground-truth'', was \emph{not} used as input to the network; it is
provided here to show the actual letter of the same font.  In the
second column, note the similarity of the correctly discerned letters
from the ground-truth.  In particular, in this column, row 1 had very
few distinguishing marks between the proposed and real 'H'; in row 4,
the 'U' was recognized as not being a member of the same font, as was
the 'V' in row 7 -- based solely on the weight of the strokes.

The third column shows the false-negatives. Several of the mistakes
are readily explained: rows 5,6,7 are non-alphabetic fonts (no
pre-filtering was done in the font selection process).  Rows 2 and 9
are explicitly designed to have a diverse set of characters,
\emph{e.g.} for creating 'Ransom Note' like artistic effects.  Also shown are
color-inverted fonts (row 4); the likely cause of the mistakes was the
sparsity of such fonts in the training set.

Finally, in the last column, false-positives are shown -- different
fonts that are mistakenly recognized as the same.  Three types of
errors were seen. When the fonts have extremely thin strokes (row 2),
the failure rate increases.  This is due to two factors:
under-representation in the training set and less information in the
pixel maps from which to make correct discriminations.  Second, as
before, non-alphabetic fonts appear.  The third source of mistakes is
that many fonts have characters that appear similar or match the basis
font's style (rows 3,7,8).  It is worth emphasizing here that the font
design process is not an algorithmic one; the artist/designer can
create fonts with as much, or as little, variation as desired.  There
may be multiple acceptable variations for each
character.

Despite the similar performance of the 7 networks, enough variation in
the outputs exists to use them as an 
ensemble~\cite{schapire1998}.  In 76.7\% of examples, all networks
were correct.  In 8.2\%, only 1 network was mistaken, in 4.1\%, 2/7
were mistaken and in 3.0\%, 3/7 were mistaken.  Therefore, by using a
simple majority voting scheme, the seven networks can be employed as
an ensemble that yields 92.1\% accuracy. This ensemble will be used
throughout the remainder of the paper.

\section {A Generative Task: Creating Style-Specific Characters}

In the previous section, we determined that there was sufficient
information in the 4-Basis letters to correctly determine whether a
fifth letter was a member of the same font.  In this section, we
attempt to construct networks that generate characters.  The
experiments are broadly divided into two approaches: single and
multiple-letter generation.  In the first approach, a network is
trained to generate only a single letter.  In the multiple-letter
generation networks, all letters are generated simultaneously by the
same network. Though the single letter networks are conceptually
simpler, training a network to generate multiple letters
simultaneously may allow the hidden units to share useful
representations and features -- \emph{i.e.} serif style, line width,
angles, etc.  This is a form of transfer learning with a strong basis
in multi-task
learning~\cite{bengio2012,huang2013cross,oquab2014learning,caruana1997multitask}.

\subsection {Single Letter Generation}

As in the previous section, experiments with numerous network
architectures were conducted.  The architectures varied in the number
of layers (2-10), units (100s-1000s) and connectivity patterns.  The
final architecture used is shown in see
Figure~\ref{singleMulti}(Left).  In the discrimination task described
in the previous section, there was a single output node.  This allowed
the use of large penultimate hidden layers since the number of
connections to the final output remained small.  In contrast, in image
generation tasks, the number of output nodes is the number of pixels
desired; in this case it is the size of a full input character
(36$\times$36).  This drastically increases the number of connections
in the network.  To alleviate this growth in the number of
connections, \emph{reverse-retinal} connections were used in which
each hidden unit in the penultimate layer connects to a small patch in
the output layer.  Various patch sizes were tried; patches of both
sizes 3 and 4 were finally employed.  Unlike other convolution and
de-convolution networks, the connection weights are \emph{not shared}.
Sharing was not necessary since translation invariance is not needed
in the generated image; each letter should be generated in the center
of the output.  Similar architectures have been used for
super-resolution and image deconvolution, often with  weight
sharing~\cite{xu2014deep,dong2014learning}.

The letter 'R' was chosen as the first test case, since, for many
fonts, the constituent building blocks for 'R' are present in the
basis letters (\emph{e.g.} copy the 'P' shape from the 'B'-basis
letter and combine it with the bottom right leg of the 'A').  As in
the previous experiments, the network was trained with the same set of
9,000 unique fonts with $L_2$ loss in pixel-space, using SGD+Momentum.
Numerous experiments, both with and without batch
normalization~\cite{ioffe2015batch}, were conducted -- no consistent
difference in final error was observed. Results for fonts from the
test set are shown in Figure~\ref{RresultsFig}.

\begin{figure}

 \tiny
\centering

 \begin {tabular} {   S| I | D | D | D| D |  D | T | D }
 \tiny

 & Inputs  & Gen. & SSE &shape & serif &  accept? & comments by rater
 & Actual\\
 \hline
 \hline

\input{arch-figs/tableR.tex}

 \end {tabular}
\caption {16 Results shown for 'R' generation.  Results are judged by
  an external human evaluator.  Actual font characters shown on the
  right.  12/16 were deemed as acceptable replacements for the original.}
\label{RresultsFig}

\end{figure}

 Unfortunately, simply measuring the pixel-wise difference to the
 actual target letter (SSE) does not correlate well to subjective
 evaluations of performance.  Yet, as in all real-world image
 generation tasks, aesthetic judgment is vital.  We asked a human
 rater\footnote{The subjective evaluation was conducted by an
   independent User Experience Researcher (UER) volunteer not
   affiliated with this project.  The UER was given a paper copy of
   the input letters, the generated letters, and the actual letter.
   The UER was asked to evaluate the 'R' along the 3 dimensions listed
   above.  Additionally, for control, the UER was also given examples
   (not shown here) which included real 'R's in order to minimize
   bias.  The UER was not paid for this experiment.} to ascertain the
 quality of the result along 3 axis: (1) is the shape generally
 correct, (2) are the serif appropriately captured, (3) is this an
 acceptable replacement for the actual letter?  The last metric is the
 hardest to measure, but perhaps the most relevant.  The outcome was
 overall positive, with the caveats noted in Figure~\ref{RresultsFig}.

\subsection {Simultaneous Letter Generation - Multi-Task Learning}

\begin{figure}[]
\centering
\includegraphics[width=0.98\textwidth]{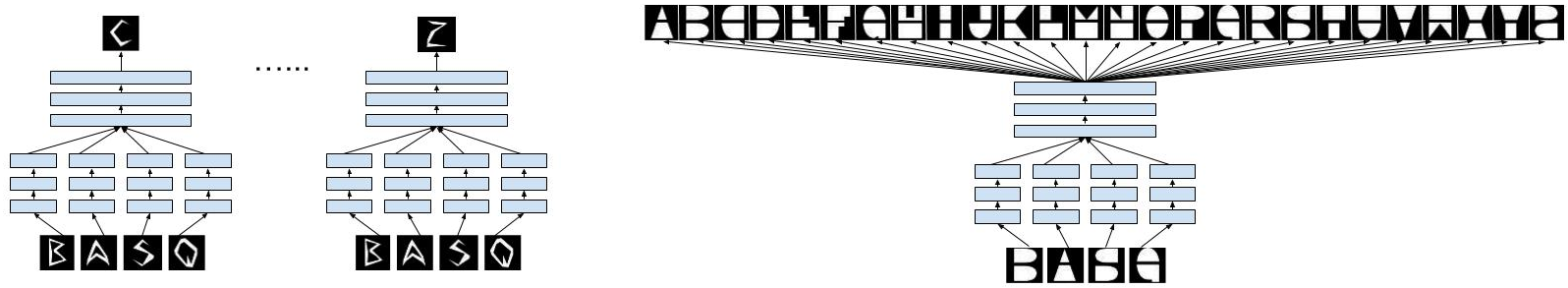}
\caption{Single Vs. Multi-Letter Generation.  The left networks show
  individual letter creation (only 2 of 22 shown).  The right network
  generates all letters simultaneously, thereby allowing the hidden
  units to share extracted information.  The tower/column
  architectures are employed to transform the basis letters (as was done 
  for the discrimination task).  The
  generative network also \emph{auto-encodes} the basis
  letters (Right).}
\label{singleMulti}
\end{figure}

In this section, we explore the task of generating all the upper-case
letters simultaneously.  Unlike the single-letter generation process
described in the previous section, the hidden units of the network can
share extracted features.  This is particularly useful in the
character generation process, as many letters have common components
(\emph{e.g.} 'P' \& 'R' , 'O' \& 'Q', 'T' \& 'I').  See
Figure~\ref{singleMulti}(Right).

\begin{figure}
\centering
\includegraphics[width=0.90\textwidth]{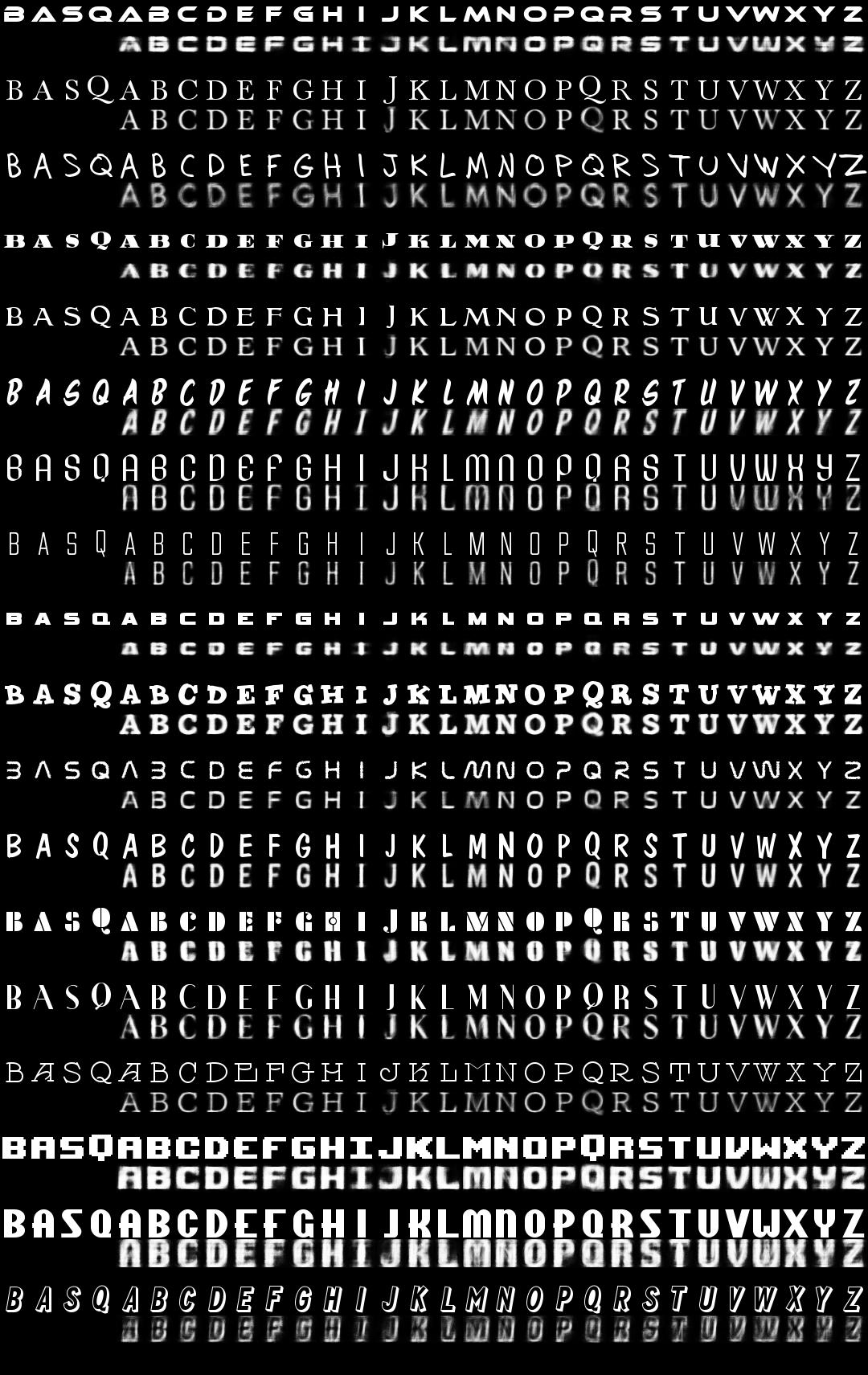}
\caption{Raw Output of the multi-letter-generation networks, with
  \textbf{no post-processing}.  18 examples shown.  Top good results.
  Bottom, results where some of the letters are smudged.
  Non-alphabet/picture fonts not shown.}
\label{multiRes}
\end{figure}

Once again, numerous architectures were explored (over 30 in total).
Comparing the best single-letter-generation networks with the best
multi-letter-generation networks, the SSE error was repeatedly reduced
(by 5\%-6\%) by using the multi-letter generation networks.
Qualitatively, however, the characters generated by both networks
appeared similar.  Nonetheless, because of the small SSE error
improvement coupled with the ease of deployment, the
multi-letter-generation network are used going forward. A large set of
results are shown in Figure~\ref{multiRes}.  Subjectively evaluating
these results would be error prone and yield inconsistent results
given the difficulty in evaluating the individual consistency of a
large set of characters with each other and with respect to the basis
set.  Instead, in the next section, we describe novel methods for
validating the results.

\subsection {Validating Results}
\label{validating}

In the previous section, the results of the font generation network
were measured by the network's SSE error and subsequently by an
external human evaluator.  Here, we present alternate methods based on
using the discriminative networks developed in
Section~\ref{discriminator}.  The discriminative networks were used to
determine whether an input character $\Phi$ was the same font as the
basis characters.  Recall that by employing a voting-ensemble of 7
networks, 92.1\% accuracy was attained.  

Here, to test the generative network $G$ on font $f$, the results of
$G(f)$ are passed into the discriminative network ensemble ($D$) along with
the original BASQ-basis letters, $D(B_f,A_f,S_f,Q_f, G(f))$.
Consequently, the output of $D$ is used to determine whether the
generated letter is the same font as the basis letters.  Novel
variants of this method, using pairs of generator/discriminator
networks have been independently proposed in~\cite{jiwoong2016}, which
is based on the architectures of generative-adversarial
models~\cite{goodfellow2014,dosovitskiy2016}.

In the most straightforward formulation, we test whether the letters
generated by $G$ appear the same as the original font. See
Table~\ref{multiperf} -- column \emph{Original Basis, Synthetic
  Test}. Formally, the test is:

\begin{mdframed}
\begin{small}
\begin{lstlisting}
For each font,  $f$, in the test set:
  generate all characters:
     $G(B_f,A_f,S_f,Q_f) \rightarrow  [A_{f'},B_{f'},C_{f'},D_{f'},... ,Z_{f'}]$ 

  test each GENERATED character:
     $D(B_f,A_f,S_f,Q_f,A_{f'})  \rightarrow [Same/Different]$
     ...
     $D(B_f,A_f,S_f,Q_f,Z_{f'})  \rightarrow [Same/Different]$
\end{lstlisting}
\end{small}
\end{mdframed}

Recall that the discriminative network ensemble, $D$, was trained with
tuples of letters $\{BasisSet,\Phi\}$ drawn from the training set of
fonts and tested on an entirely separate set of fonts.  Thus, we are
not limited in the set of fonts that we can use as the basis-set.  In
the next examination, we reverse the question asked above.  We
generate the letters as before, but ask the question, if the
\underline{generated} letters are used as the basis set, will the
\underline{original} letters be recognized as the same font? See
Table~\ref{multiperf} -- column \emph{Synthetic Basis, Original Test}.
The generation portion remains the same, but the test is revised to:

\begin{mdframed}
\begin{small}
\begin{lstlisting}
  test each ORIGINAL character:
     $D(B_{f'},A_{f'},S_{f'},Q_{f'},A_{f})  \rightarrow  [Same/Different]$
     ...
     $D(B_{f'},A_{f'},S_{f'},Q_{f'},Z_{f})  \rightarrow  [Same/Different]$
\end{lstlisting}
\end{small}
\end{mdframed}

In the previous two experiments, we compared the generated characters
to the original font.  This gives a measure of how closely the
generated characters resemble the original font.  In the final
experiment, we ask how consistent the generated letters are \emph{with
  each other}.  If we are given the basis letters from the generated
font, will the other generated letters from the same font be
classified as the same?  See Table~\ref{multiperf} -- column
\emph{Synthetic Basis, Synthetic Test}.

\begin{mdframed}
\begin{small}
\begin{lstlisting}
  test GENERATED characters against GENERATED basis:
     $D(B_{f'},A_{f'},S_{f'},Q_{f'},A_{f'})  \rightarrow  [Same/Different]$
     ...
     $D(B_{f'},A_{f'},S_{f'},Q_{f'},Z_{f'})  \rightarrow  [Same/Different]$
\end{lstlisting}
\end{small}
\end{mdframed}

Looking at Table~\ref{multiperf}, from the column~\emph{Original
  Basis, Synthetic Test}, it can be seen that the generated fonts
appear similar to the original fonts on which they are based.  An
interesting question arises: why, then, does the second
test~\emph{Synthetic Basis, Original Test} have a lower recognition
rate?  The answer lies in the fact that the font designer may
introduce non-systematic variations that add visual interest, but may
not adhere to strict stylistic cues present in other letters.  In the
most obvious cases, the 'ransom note fonts', each character is
intentionally not representative of other characters.  Other themed
fonts introduce objects such as trees, animals, and vehicles into the
design that reflect each artist's individual creativity.  For example,
some fonts replace the empty spaces in the letter 'B' with
mini-airplanes, but may not do so in the letter 'A', etc.  The
generated fonts, when they capture such shapes, will synthesize more
consistent characters that reuse many of the same elements.  It is
likely that the original font's glyphs may appear outside the more
cohesive set generated by the networks.\footnote{For completeness, we
  also analyzed the 'R's generated by the one-letter-at-a-time
  networks.  They had similar performance (when measured with $D$) to
  the 'R' row shown in Table~\ref{multiperf}, with (6\%) higher SSE.}

\begin{table}
 \centering
 \caption {Performance by Letter: (SSE \& Discriminative Network Ensemble} 
\scriptsize
 \label {multiperf}
 \begin {tabular} {  c | c |c | c | c || c}

 & & \multicolumn{4}{c}{Discriminative Network Ensemble Based Evaluation} \\
 \hline
 letter & SSE & \makecell {Original Basis\\Synthetic Test} &  \makecell {Synthetic Basis \\ Original Test} & \makecell {Synthetic Basis \\ Synthetic Test} & \makecell {Baseline:\\ Original Basis \\ Original Test}\\
 \hline
 \hline
C & 1851.0 & 99\%  & 90\% & 100\%      & 96\% \\
D & 1987.7 & 99\%  & 90\% & 100\%      & 97\% \\
E & 2006.9 & 99\%  & 89\% & 100\%      & 95\% \\
F & 1943.0 & 98\%  & 84\% & 100\%      & 93\% \\
G & 2096.7 & 99\%  & 89\% & 100\%      & 95\% \\
H & 2066.6 & 99\%  & 89\% & 100\%      & 96\% \\
I & 1495.3 & 94\%  & 83\% & 99\%       & 90\% \\
J & 1968.8 & 92\%  & 84\% & 97\%       & 91\% \\
K & 2146.3 & 99\%  & 88\% & 100\%      & 95\% \\
L & 1799.7 & 99\%  & 88\% & 100\%      & 94\% \\
M & 2697.1 & 97\%  & 83\% & 99\%       & 92\% \\
N & 2162.8 & 98\%  & 87\% & 99\%       & 94\% \\
O & 1828.5 & 98\%  & 89\% & 99\%       & 95\% \\
P & 1936.3 & 99\%  & 89\% & 100\%      & 95\% \\
R & 2135.6 & 99\%  & 90\% & 100\%      & 97\% \\
T & 1714.8 & 97\%  & 85\% & 99\%       & 94\% \\
U & 1910.5 & 99\%  & 89\% & 99\%       & 96\% \\
V & 1950.6 & 97\%  & 85\% & 99\%       & 93\% \\
W & 2611.0 & 97\%  & 82\% & 99\%       & 92\% \\
X & 1985.5 & 99\%  & 86\% & 100\%      & 94\% \\
Y & 1998.3 & 98\%  & 85\% & 99\%       & 92\% \\
Z & 1901.4 & 97\%  & 84\% & 99\%       & 91\% \\
  \hline
  \hline
\multicolumn{2}{c}{Average} &
  97\% & 86\% & 98\% & 94\% \\
 \end {tabular}

\end{table}

The third column,~\emph{Synthetic Basis, Synthetic Test}, shows that
the characters generated are extremely consistent with each other.
Normally, this would be considered a good attribute; however, when
viewed in terms of the baseline (Column 4: ~\emph{Original Basis,
  Original Test}), it raises an important question.  Why are the
generated fonts \emph{more} homogeneous than the original fonts?  Is
it for the reason mentioned above, or is it for the, potentially more
troubling, reason that all the generated characters (across all fonts)
are too much alike?  Have all the characters ``regressed to the
mean?''  As a final test, we examine this possibility explicitly using
\underline{only network generated} fonts.  First, we set the basis
letters (BASQ) from a randomly selected generated font.  Second, we
randomly select another generated font and character as the
candidate test character.  50,000 randomly paired samples are created.
Unlike the previous tests, in which accuracy was measured by how many
matches were found, the accuracy is measured by how many non-matches
are detected.  This explicitly tests whether the generated fonts look
too similar to each other.  Running this test with the discriminative
network ensemble, $D$, yields a different-font detection rate of
90.7\%. For a baseline, we repeat this experiment with 50,000 pairs
generated from the original (non-synthesized) fonts.  $D$ yields a
correct different-font detection rate of 90.1\%.  The very close
results indicate that the generated fonts \emph{have not} regressed to
the mean, in terms of recognizable serifs and style, and remain as
distinguishable as the original fonts. The difference in performance
between column 3 and 4 in Table~\ref{multiperf} is likely due to the
variability introduced by smaller artistic variances inserted by the
designers, as described in the paragraph above.

\section{Future Work}

Beyond the straightforward explorations of varying the number and
selection of basis letters and also generating lower-case letters,
there are many conceptually interesting avenues for future work.
First, an alternative approach to using the generator networks is to
use discriminative-networks and propagate derivatives back to the
inputs to modify the input pixels to maximize the similarity of the
hidden states to specified values.  This has recently been proposed to
create natural images as well as dream like images ``Deep
Dreams''~\cite{Mordvinstev2015,gatys2015}. %

Second, we made a concerted effort to use the simplest networks
possible.  An extensive empirical search through the space of networks
and learning algorithms was conducted to find the most straightforward
approach to addressing this task.  Recent
preprints~\cite{upchurch2016,jiwoong2016} describe concurrent
explorations of similar and related problems with much more complex
architectures that yield comparably promising results.  Extending this
work to other architectures is easily done; for example, the
evaluation mechanisms used in this study are akin to Generative
Adversarial Nets
(GAN)~\cite{goodfellow2014,denton2015,dosovitskiy2016} in which a
synthetic-vs.-real distinguisher network and a font generator are
trained to outperform each other.  This study provides a strong, and
easily implemented, baseline to which new architectures and learning
approaches can be compared.

\begin{figure}
\label{combinations}
\centering
    \includegraphics[width=1.0\textwidth]{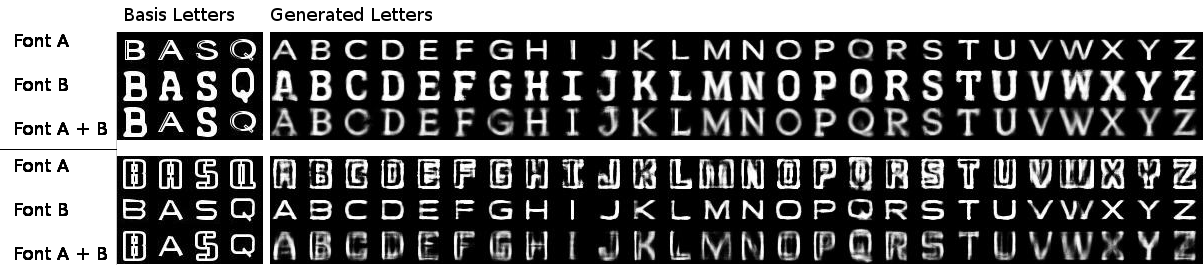}
\caption{Novel combinations of two fonts.  In each of the two results,
  the first row shows the 4 input letters and characters generated
  from $G_{fontA}$, second row for $G_{fontB}$.  In the Row 3, two
  characters from each font are used as input and the resulting
  characters shown.  Top Example: Note that the weight (thickness of
  lines) from the first font are combined with the size of the second
  font.  Bottom Example: Note that the ``hollow'' look of the first
  font is combined with the weight and shapes of the second font.}
\label{futureTease}
\end{figure}

Third, the trained networks can be used in novel manners beyond
synthesizing characters from existing fonts.  Will it be possible to
take the attributes of multiple fonts and combine them?  For example,
if the the basis set is composed of 2 characters from font-A and 2
characters from font-B, will the resulting characters be a combination
of the two?  Preliminary evidence (see Figure~\ref{futureTease})
suggests that it may be possible; though perhaps more than 2 examples
from each font will be necessary for artistically more innovative
combinations to be produced.

Fourth, we have taken an image based approach to font generation.  An
alternate, more speculative direction, is to use a non-image based
approach with recurrent networks, such as Long-Short-Term-Memory
(LSTMs).  Can LSTMs be used to generate characters \emph{directly} in
TrueType language?  For this task, network's inputs would include the
TTF of the basis letters.  Similar programmatic learning in which
LSTMs compose simple programs and sequences of music have been
recently
attempted~\cite{zaremba2014learning,eck2002first,graves2014neural}.
If it is possible to generate results directly in TTF encodings, this
will produce another, orthogonal, result to complement this study.
The feasibility of this approach is open for future research.

\section{Conclusions}

In this paper, we have presented a learning method to analyze
typographic style based on a small set of letters and to employ the
learned models to both distinguish fonts and produce characters in the
same \emph{style.}  The results are quite promising, but fully
capturing artistic visual intent is just beginning.  Many of the
overall shapes, weights, angles and serifs were successfully modeled.
In the future, as learning progresses, some of the more individualized
nuances and repeated intricate design patterns, unique to individual
fonts, will be captured.

 \section{Acknowledgments}

\begin{figure}
\centering

    \fbox {
    \includegraphics[width=0.9\textwidth]{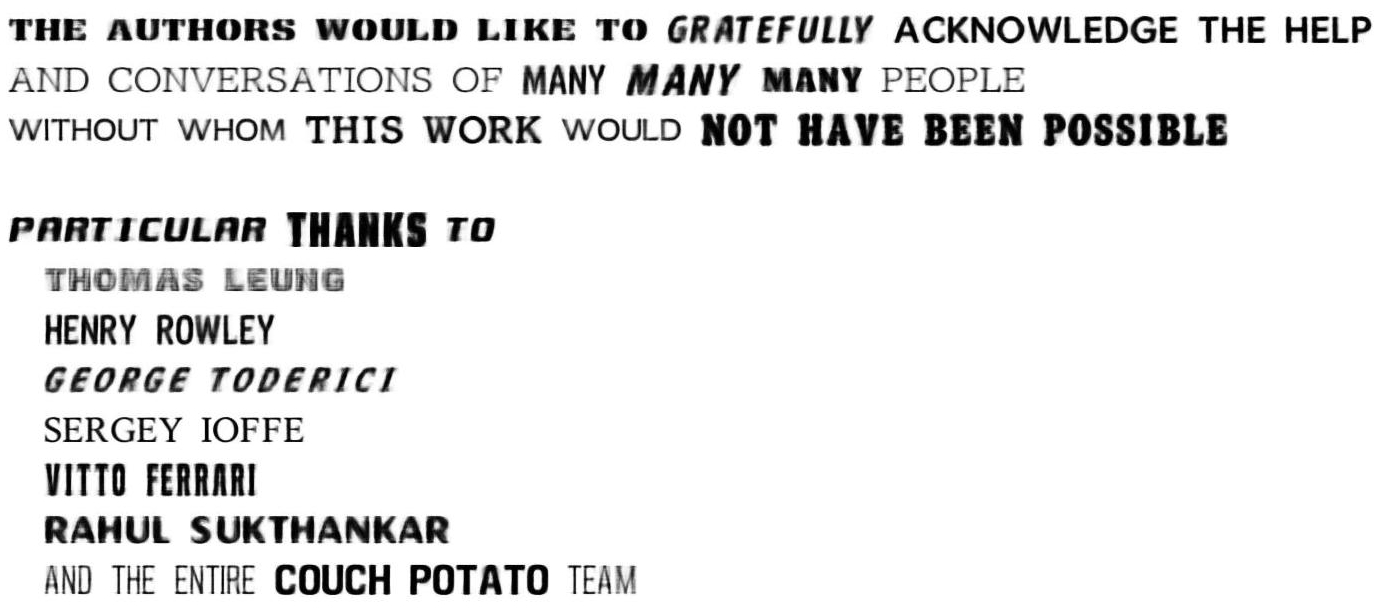}
}
\caption{Acknowledgments displaying 24 synthesized fonts (from test set
  only).}
\label{acks}
\end{figure}

\bibliographystyle{splncs}
\bibliography{fonts.bib}

\end{document}